%% file: main.tex
\documentclass[runningheads]{llncs}

\usepackage{eccv}

\usepackage{eccvabbrv}

\usepackage{graphicx}
\usepackage{booktabs}

\usepackage[accsupp]{axessibility}  

\usepackage{marvosym}  
\usepackage[dvipsnames]{xcolor}
\usepackage{arydshln}
\usepackage{multirow}

\usepackage{wrapfig}
\usepackage{amsmath}

\usepackage{listings}
\definecolor{codegreen}{rgb}{0,0.6,0}
\definecolor{codegray}{rgb}{0.5,0.5,0.5}
\definecolor{codepurple}{rgb}{0.58,0,0.82}
\definecolor{backcolour}{rgb}{0.95,0.95,0.92}
\lstdefinestyle{mystyle}{
    backgroundcolor=\color{backcolour},   
    commentstyle=\color{codepurple},
    keywordstyle=\color{magenta},
    numberstyle=\tiny\color{codegray},
    stringstyle=\color{codegreen},
    basicstyle=\ttfamily\scriptsize,
    breakatwhitespace=true,         
    breaklines=true,
    captionpos=b,
    keepspaces=true,
    numbers=left,
    numbersep=5pt,
    showspaces=false,
    showstringspaces=false,
    showtabs=false,
    tabsize=2
}
\usepackage[pagebackref,breaklinks,colorlinks,citecolor=eccvblue]{hyperref}

\usepackage{orcidlink}

\begin{document}

\title{OpenPSG: Open-set Panoptic Scene Graph Generation via Large Multimodal Models}

\titlerunning{OpenPSG}


\author{
  Zijian Zhou${}^{1}$ \quad 
  Zheng Zhu${}^{2}$ \quad 
  Holger Caesar${}^{3}$ \quad  
  Miaojing Shi${}^{4,5}$\textsuperscript{\Letter} \quad \\
}

\authorrunning{Z.~Zhou, Z.~Zhu, H.~Caesar, M.~Shi}

\institute{
  ${}^{1}$Department of Informatics, King's College London \quad
  ${}^{2}$GigaAI\\
  ${}^{3}$Intelligent Vehicles Lab, Delft University of Technology\\
  ${}^{4}$College of Electronic and Information Engineering, Tongji University\\
  ${}^{5}$Shanghai Institute of Intelligent Science and Technology, Tongji University\\
}

\maketitle
\footnotetext{\textsuperscript{\Letter} Corresponding author.}
\input{sections/0_abstract}
\input{sections/1_introduction}
\input{sections/2_related_work}
\input{sections/3_method}
\input{sections/4_experiments}
\input{sections/5_conclusion}

\newpage

\section*{Acknowledgments}
The authors would like to thank Prof. Tomasz Radzik for helpful discussions.
Computing resources provided by King’s Computational Research, Engineering and Technology Environment (CREATE).
This work was supported by the Fundamental Research Funds for the Central Universities.

\bibliographystyle{splncs04}
\bibliography{main}

\newpage
\input{sections/6_appendix}


\end{document}

%% file: sections/0_abstract.tex
\begin{abstract}
Panoptic Scene Graph Generation (PSG) aims to segment objects and recognize their relations, enabling the structured understanding of an image.
Previous methods focus on predicting predefined object and relation categories, hence limiting their applications in the open world scenarios.
With the rapid development of large multimodal models (LMMs), significant progress has been made in open-set object detection and segmentation, yet open-set relation prediction in PSG remains unexplored.
In this paper, we focus on the task of open-set relation prediction integrated with a pretrained open-set panoptic segmentation model to achieve true open-set panoptic scene graph generation (\textbf{OpenPSG}).
Our OpenPSG leverages LMMs to achieve open-set relation prediction in an autoregressive manner.
We introduce a relation query transformer to efficiently extract visual features of object pairs and estimate the existence of relations between them. The latter can enhance the prediction efficiency by filtering irrelevant pairs.
Finally, we design the generation and judgement instructions to perform open-set relation prediction in PSG autoregressively.
To our knowledge, we are the first to propose the open-set PSG task.
Extensive experiments demonstrate that our method achieves state-of-the-art performance in open-set relation prediction and panoptic scene graph generation.
Code is available at \url{https://github.com/franciszzj/OpenPSG}.

\vspace{-2mm}
\keywords{Panoptic Scene Graph Generation \and Open-set \and Large Multimodal Models}

\end{abstract}

%% file: sections/1_introduction.tex
\section{Introduction}
\label{sec:introduction}

\begin{figure}[h]
    \centering
    \includegraphics[width=1.0\linewidth]{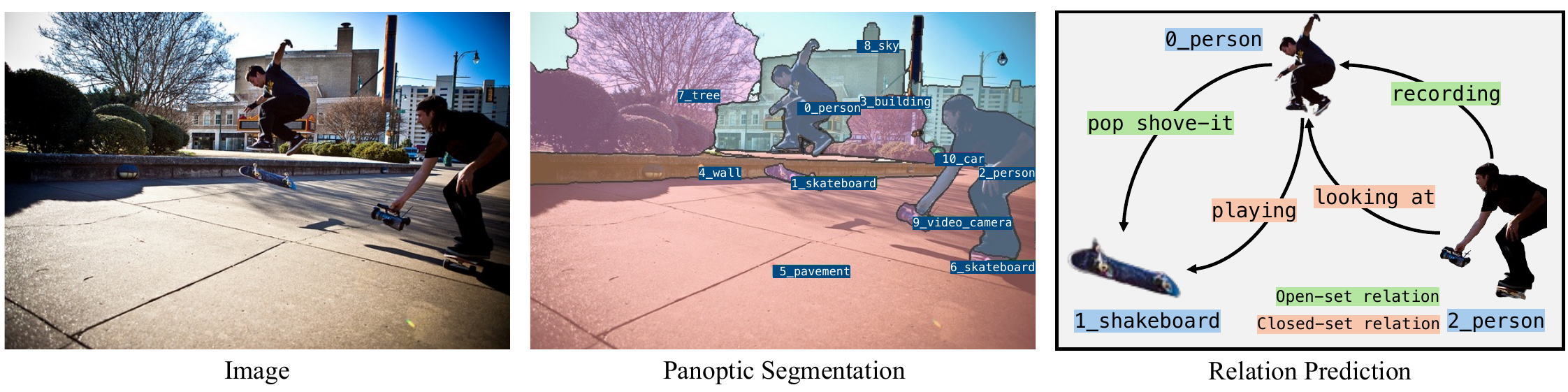}
    \caption{The left image is the input to our OpenPSG, the middle one displays the panoptic segmentation result, and the right one shows the predicted relations between objects.
    Our method can predict both known (close-set) relations, \eg, (0\_person, \textit{playing}, 1\_skateboard), (2\_person, \textit{looking at}, 0\_person), and unknown (open-set) relations, \eg, (0\_person, \textit{pop shove-it}, 1\_skateboard), (2\_person, \textit{recording}, 0\_person).}
    \label{fig:open_figure}
\end{figure}

Panoptic scene graph generation (PSG)~\cite{yang2022panoptic} aims to segment objects within an image and recognize the relations among them, thereby constructing a panoptic scene graph for a structured understanding of the image.
Given its significant potential in applications such as visual question answering~\cite{hildebrandt2020scene}, image captioning~\cite{gao2018image, chen2020say}, and embodied navigation~\cite{singh2023scene}, PSG has attracted considerable attentions from researchers ever since its emerging~\cite{zhou2023hilo, wang2023pair, li2023panoptic, yang2023focusing, zhao2023textpsg, zhou2023vlprompt}.

Previous PSG methods \cite{zhou2023hilo, wang2023pair, zhou2023vlprompt} are only capable of predicting closed-set object and relation categories while failing to recognize objects/relations beyond predefined categories.
Recently, with the advent of large multimodal models (LMMs) such as CLIP~\cite{radford2021learning}, BLIP-2~\cite{li2023blip} \etc, a significant number of open-set prediction methods for object detection~\cite{du2022learning, zang2022open, lin2022learning, zhang2023simple, yao2023detclipv2} and segmentation~\cite{ghiasi2022scaling, liang2023open, zhang2023simple, yu2023towards} are introduced, attributing to LMMs' rich understanding of language and strong connections between vision and language.
Nevertheless, open-set prediction of relations has been largely unexplored so far.

Compared to open-set object detection and segmentation, open-set relation prediction is more complex: 
the model is required to both understand different objects and  recognize relations of object pairs based on their interactions; especially,  the computation of the latter can be exponentially increased.  
To bridge the gap, in this paper, we focus on the open-set relation prediction. 

LLMs~\cite{li2023blip, liu2024visual, zhu2023minigpt} have demonstrated exceptional semantic analysis and understanding abilities across various multimodal tasks.
In particular with the text processing, LMMs are not only good at interpreting on nouns (\ie  representing objects) but also pay considerable attention on predicates (\ie representing relations between objects), ensuring their generated contents to be sufficiently coherent~\cite{achiam2023gpt}.
Inspired by this, we propose the \textbf{Open}-set \textbf{P}anoptic \textbf{S}cene \textbf{G}raph Generation architecture, \textbf{OpenPSG}, leveraging the capabilities of LMMs for open-set relation prediction.

To this end, we utilize a large multimodal model (\eg, BLIP-2~\cite{li2023blip}) to achieve open-set relation prediction.
Specifically, our model comprises three parts.
First, the \textit{open-set panoptic segmenter}, we adapt an existing model (\eg, OpenSeeD~\cite{zhang2023simple}) which is capable of extracting open-set object categories, masks, and visual features from the whole image, forming object pairs and pair masks.
Second, the \textit{relation query transformer}, which has two functions: extracting visual features of object pairs based on pair masks and with a special focus on pair interactions; judging the potential relations between object pairs.
They are realized by two sets of queries, pair feature extraction query and relation existence estimation query.
Only those object pairs that are judged to likely have relations are fed into the third part, the \textit{multimodal relation decoder}.
This decoder directly inherits from the LMM to predict the open-set relations given an object pair in an auto-regressive manner, on condition of specifically-designed text instructions and pre-extracted pair visual features.

To the best of our knowledge, we are the first to propose the task of open-set panoptic scene graph generation, enabling the open-set prediction of both object masks and relations. 
Extensive experiments demonstrate that our OpenPSG achieves state-of-the-art results in the closed-set setting and  exhibits outstanding performance in the open-set setting.

%% file: sections/2_related_work.tex
\section{Related Work}
\label{sec:related_work}

\subsection{Panoptic Scene Graph Generation}
Panoptic scene graph generation stems from scene graph generation (SGG) by replacing bounding boxes with panoptic segmentation masks to represent the objects, so as to achieve a more comprehensive understanding of scenes.
Following the introduction of PSG~\cite{yang2022panoptic}, a series of related works~\cite{yang2022panoptic, wang2023pair, zhou2023hilo, li2023panoptic, zhao2023textpsg, zhou2023vlprompt} emerge, significantly advancing its performance.
For example, Yang~\etal~\cite{yang2022panoptic}, based on DETR~\cite{carion2020end}, introduce an end-to-end framework with learnable queries to generate panoptic scene graphs.
Wang~\etal~\cite{wang2023pair} design a pair proposal network to filter irrelative subject-object pairs, achieving performance improvement of the PSG.
Subsequently, Zhou~\etal~\cite{zhou2023hilo} build a HiLo architecture based on Mask2Former~\cite{cheng2022masked} and devise separate branches for high- and low-frequency relations respectively, hence achieving an unbiased relation prediction method.
Li~\etal~\cite{li2023panoptic} re-balance the relation prediction  by adaptively transferring information from high-frequency to low-frequency relations.
Additionally, Zhao~\etal~\cite{zhao2023textpsg} implement a weakly-supervised PSG method given only image-text pairs as annotations, allowing for the learning of panoptic scene graphs from image-level ground truth.
Recently, Zhou~\etal~\cite{zhou2023vlprompt} leverage the rich language information inherent in large language models~\cite{achiam2023gpt} and design effective image-text interaction modules to assist unbiased PSG.
All these works are learned in the closed-set setting, in this paper, we study the open-set PSG, which has been unexplored.

\subsection{Open-set Scene Graph Generation}
Open-set SGG has been studied in recent years. 
Earlier works~\cite{kan2021zero, yu2022zero}, often termed as zero-shot SGG, focus on transferring the knowledge from known relations to unknown relations given their prior connections; for example, Kan~\etal~\cite{kan2021zero} leverage the external commonsense knowledge while Yu~\etal~\cite{yu2022zero} use knowledge graphs for the knowledge transfer.
Subsequently, with the advancement of large multimodal models, various open-set SGG works~\cite{he2022towards, zhang2023learning, yu2023visually, chen2023expanding} have emerged; for example, He~\etal~\cite{he2022towards} and Zhang~\etal~\cite{zhang2023learning} focus on predicting relations between unknown objects in the SGG using multimodal models.
Yu~\etal~\cite{yu2023visually} and Chen~\etal~\cite{chen2023expanding} on the other hand focus on the open-set relation prediction in PSG, sharing the same aim with us.
Specifically, Yu~\etal~\cite{yu2023visually} utilize the CLIP model to match visual features with textual relation features for open-set relation prediction; Chen~\etal~\cite{chen2023expanding} employ a student-teacher network to align visual concepts in multimodal models for predicting open-set relations.
In this paper, different from previous works, we introduce an auto-regressive method based on the LMM to achieve open-set relation prediction.

\subsection{Large Multimodal Models}
Since the introduction of large models like the GPT series~\cite{achiam2023gpt}, recent years have witnessed rapid development in large multimodal models~\cite{radford2021learning, li2023blip, liu2024visual}.
Benefiting from its nature of connecting vision and language, LMMs have significantly advanced various downstream tasks, ranging from computer vision~\cite{du2022learning, yu2023towards} to natural language processing~\cite{wang2020language, joshi2019bert, raffel2020exploring}. 
Early multimodal models, such as CLIP~\cite{radford2021learning}, trained on image-text paired datasets through contrastive learning to align the visual and textual information.
Subsequently, owing to enlightenment of the autoregressive prediction in large language models~\cite{achiam2023gpt, touvron2023llama}, there has been an explosive growth of LMMs~\cite{li2023blip, liu2023improved, liu2024visual}; by introducing mechanisms that can transform visual information into large language models, they facilitate the communication between visual and textual information.
Furthermore, this has endowed LMMs with the capability to generate free text, leading to substantial improvements in numerous multimodal tasks.
In this paper, we leverage LMMs to design a multimodal relation decoder to predict relations in an open-set scenarios.

%% file: sections/3_method.tex
\section{Task Definition}
\label{sec:task_definition}

We define the task, open-set panoptic scene graph generation.
Given an image $I \in \mathbb{R}^{H \times W \times 3}$, the objective of this task is to extract an open-set panoptic scene graph $G = \{O, R\}$ from the image $I$, where $H$ and $W$ are the height and width of the image.
Here:
\begin{itemize}
    \item $O = \{o_i\}_{i=1}^{N}$ represents $N$ objects segmented from the image, each defined as $o_i = \{c, m\}$, where $c$ is the object category that can belong to either predefined base object categories $C_{base}$ or undefined novel object categories $C_{novel}$.
    $m$ represents the binary mask in $\{0, 1\}^{H \times W}$ of the object.
    \item $R = \{r_{i,j} \mid i,j \in \{1, 2, \ldots, N\}, i \neq j\}$ represents the relations between objects, where $r_{i,j}$ denotes the relation between $o_i$ and $o_j$, with $o_i$ as the subject and $o_j$ as the object.
    Each relation $r$ can belong to either predefined base relation categories $K_{base}$ or undefined novel relation categories $K_{novel}$.
\end{itemize}

\begin{figure}[t]
    \centering
    \includegraphics[width=1.0\linewidth]{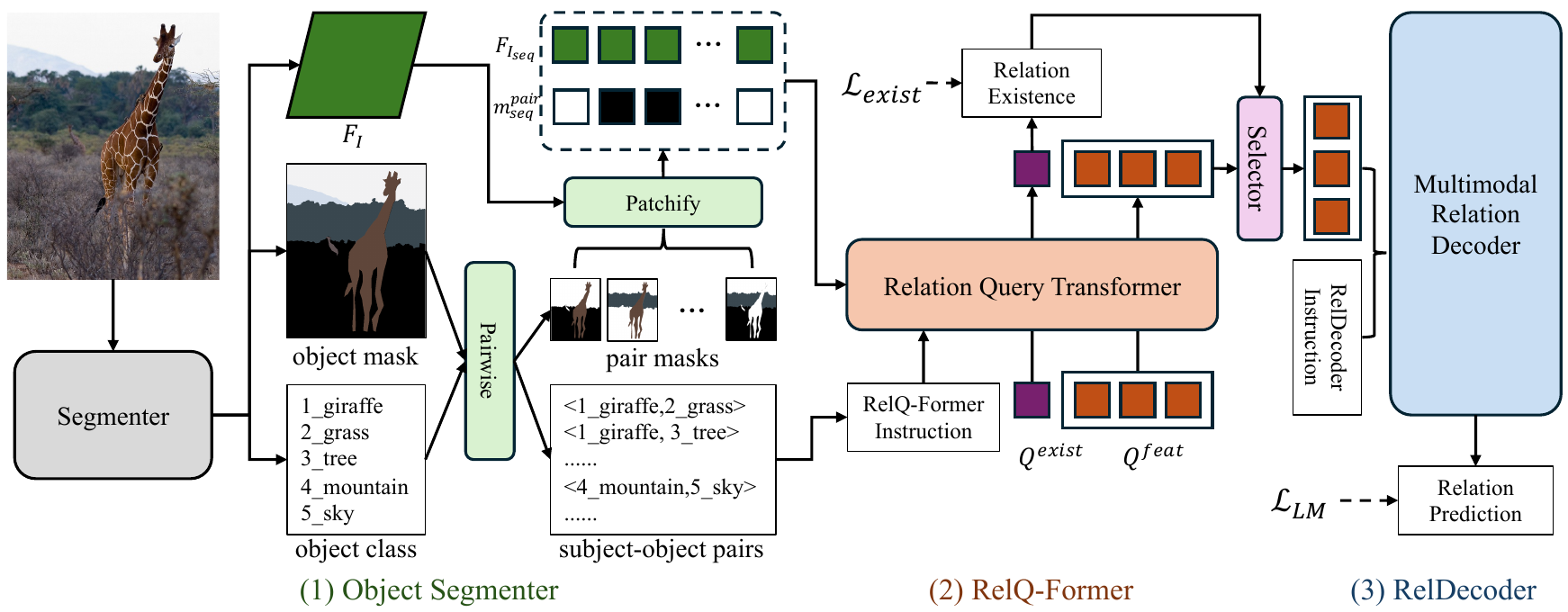}
    \caption{The overall framework of our {OpenPSG}, which comprises three components: object segmenter, relation query transformer and multimodal relation decoder.}
    \label{fig:overview}
\end{figure}

\section{Method}
\label{sec:method}

As illustrated in Fig.~\ref{fig:overview}, our {OpenPSG} comprises three components: object segmenter, relation query transformer (RelQ-Former), and multimodal relation decoder (RelDecoder).
For the object segmenter (Sec.~\ref{sec:object_segmenter}), we utilize a pretrained open-set panoptic segmentation model to transform the input image into object categories and masks, as well as visual feature representing the whole image.
Subsequently, we input the object categories, masks, and visual feature into the RelQ-Former (Sec.~\ref{sec:rel_q_former}). Through two sets of learnable queries and complemented by designed instructions, we obtain visual features of object pairs compatible with LMMs' input format as well as judgements on potential relation existence.
Finally, only those object pairs being judged to likely have a relation are sent into the RelDecoder (Sec.~\ref{sec:rel_decoder}) for open-set relation prediction, ultimately yielding an open-set panoptic scene graph.

\subsection{Object Segmenter}
\label{sec:object_segmenter}
Given an image $I$, we utilize the pretrained open-set object segmenter (\eg, OpenSeeD~\cite{zhang2023simple}) to predict the objects $O$ within the image and the whole-image visual feature $F_I \in \mathbb{R}^{h \times w \times D}$. 
Here, $h$ and $w$ represent the height and width of $F_I$, and $D$ denotes the feature dimension.
The segmenter has a similar architecture to Mask2Former~\cite{cheng2022masked} includeing a pixel decoder.
The whole-image visual feature $F_I$ refers to the visual feature output by the pixel decoder.
Below, we develop the patchify module and pairwise module to process the output of the segmenter, generating the input for RelQ-Former. 

\subsubsection{Patchify Module.}
Patchify module aims to serialize visual feature $F_I$ and object masks $m$, enabling them to be processed as inputs by the RelQ-Former (Sec.~\ref{sec:rel_q_former}).
Similar to the input patchify layer of vision transformer (ViT)~\cite{dosovitskiy2020image}, we utilize a single convolution layer to transform the extracted $F_I$ into a sequence of visual tokens $F_{Iseq} \in \mathbb{R}^{L \times D}$, where $L$ is the number of patches and $D$ is the feature dimension.
When the kernel size and stride of the convolution layer are both $p$, $L$ is calculated as $L = \frac{h}{p} \times \frac{w}{p}$.
Simultaneously, we employ nearest neighbor interpolation to each extracted object's mask $m_i$, where the size of $m_i$ is of height $\frac{h}{p}$ and width $\frac{w}{p}$, and then reshape it to a one-dimensional vector with the length $L$.
After processing all masks in the same way, we obtain the mask sequence $m_{seq} \in \{0, 1\}^{N \times L}$ for all objects.

\subsubsection{Pairwise Module.}
Pairwise module aims to construct subject-object pairs.
Given $N$ objects in the image $I$, we pairwise all objects $O$ into subject-object pairs $P = \{(o_i, o_j) | i, j \in \{1, 2, \ldots, N\}, i \neq j\}$.
The number of subject-object pairs in $P$ is $N \times (N - 1)$, which exhibits exponential growth as $N$ increases.
Consequently, we also obtain the combined subject-object pair category set $c^{pair} \in \{(c_i, c_j) | i, j \in \{1, 2, \ldots, N\}, i \neq j\}$.
We construct the mask sequences for the two objects corresponding to the indices $i$ and $j$ from $m_{seq}$ for each subject-object pair by using the logical \emph{OR} operation.
This operation is performed for all subject-object pairs, resulting in the pair mask sequence $m_{seq}^{pair} \in \{0, 1\}^{N \times (N-1) \times L}$ for the subject-object pairs, where $L$ is the number of patches.

\subsection{Relation Query Transformer}
\label{sec:rel_q_former}
Relation query transformer, leveraging the obtained $F_{Iseq}$, $c^{pair}$, and $m_{seq}^{pair}$, employs two distinct types of queries, pair feature extraction query and relation existence estimation query, along with customized instructions.
This approach facilitates the extraction of subject-object pair features and assesses which subject-object pairs likely have relations.

\subsubsection{Pair Feature Extraction Query.}
The objective of the pair feature extraction query is to extract corresponding subject-object pair features from the whole image visual feature based on the subject-object pair masks.
A common extraction method involves mask pooling~\cite{dai2015convolutional}, which extracts features for the target subject-object pair, treating each area on the subject-object pair equally.
However, for features used in relation prediction, they should focus more on areas where interactions between objects occur. 
By leveraging attention mechanisms, we facilitate interactions among visual tokens representing different areas within the visual feature sequence $F_{Iseq}$ of a subject-object pair.
This way can enhance areas that are crucial for relation predictions.
Furthermore, inspired by \cite{liu2023improved}, we design an instruction to assist the this learnable query in understanding its purpose for extracting subject-object pair features. 

Specifically, for each subject-object pair $(o_i, o_j)$, we first input the pair feature extraction query $Q^{feat} \in \mathbb{R}^{E \times D}$ into a self-attention layer ($SA(\cdot)$), along with the pair instruction designed specifically for the pair feature extraction query.
This pair instruction is processed through a tokenizer layer to obtain $F_{Inst}^{feat} \in \mathbb{R}^{X^{feat} \times D}$, which specifies the function of the pair feature extraction query, namely ``Extracting subject-object ($c_i$, $c_j$) features from visual features according to the mask''.
Here $E$ is the token number of the pair feature extraction query, and $X^{feat}$ is the token number of the pair instruction.
Note that we also incorporate the category names of the subject and object $(c_i, c_j)$ into this pair instruction.
This operation is formulated as
\begin{equation}
    F_{SA}^{feat} = Trunc(SA(Concat(Q^{feat}, F_{Inst}^{feat})), E),
\end{equation}
where $Concat(\cdot)$ denotes the concatenation operation, $Trunc(\cdot)$ represents the truncation operation, and $E$ in this truncation operation indicates that we only extract the first $E$ features, namely the features corresponding to the pair feature extraction query.
Next, we use a mask cross-attention layer ($MaskCA(\cdot)$), with $F_{SA}^{feat}$ as the query, $F_{Iseq}$ as key and value, and $m_{seq}$ as the mask, to extract features corresponding to the subject-object pair, formulated as
\begin{equation}
    F_{CA}^{feat} = MaskCA(F_{SA}^{feat}, F_{Iseq}, m_{seq}).
\end{equation}
The features $F_{CA}^{feat}$ are further refined through a feed-forward network ($FFN(\cdot)$), formulated as
$F_{FFN}^{feat} = FFN(F_{CA}^{feat})$.

By repeating this process twice, we obtain the visual features for the subject-object pair to be input into the multimodal relation decoder, $F_{I}^{pair(i,j)} \in \mathbb{R}^{E \times D}$.
We perform these operations in parallel for all subject-object pairs to obtain the corresponding features for all pairs.

\subsubsection{Relation Existence Estimation Query.}
In addition to the pair feature extraction query, we also design a relation existence estimation query to determine whether a relation likely exists between the subject $o_i$ and object $o_j$, without predicting the specific relation category.
The objective is to filter out irrelevant subject-object pairs to save the computation for subsequent LMM decoding.

Specifically, for each subject-object pair $(o_i, o_j)$, the relation existence estimation query $Q^{exist} \in \mathbb{R}^{1 \times D}$, similar to the pair feature extraction query, is input into the self-attention, mask cross-attention, and feed-forward network layers, interacting respectively with $F_{Iseq}$, $m_{seq}$ and the specially designed relation instruction.
The purpose of the relation instruction is to direct the relation existence estimation query towards determining whether a relation likely exists in the subject-object pair, \eg ``Is there a relation between $o_i$ and $o_j$?''
The relation instruction, after being processed by the tokenizer, results in $F_{Inst}^{exist} \in \mathbb{R}^{X^{exist} \times D}$, where $X^{exist}$ represents the number of tokens.
Eventually, the extracted features are input into a relation existence prediction layer, which includes a 2-layer MLP, and the predicted scores are normalized to $[0, 1]$ using the sigmoid function.
It is worth noting that we train it using binary labels indicating whether a relation exists between objects, and a selector module specified below is utilized to perform filtering during inference.

\textbf{Selector.}
The selector module implemented by 2-layer MLP is set to filter irrelvant subject-object pairs.
Only those with a score higher than the threshold $\theta$ can be input into the multimodal relation decoder.
Compared to predicting for all subject-object pairs, this can enables a 20$\times$ speedup in our experiment.

\subsection{Multimodal Relation Decoder}
\label{sec:rel_decoder}
Multimodal relation decoder aims to utilize the subject-object pair feature $F_I^{i, j}$ extracted by the aforementioned modules, combined with an instruction guiding it to achieve open-set relation prediction.
Inspired by \cite{yue2023object, yu2023towards}, we first design a \textit{generation instruction} to perform open-set relation prediction in an autoregressive manner.
This works well, yet we find that it tends to favor common relations more or less.
Therefore, we further design a \textit{judgement instruction}, leveraging the LMMs' strong analytical and judgment capabilities.
The judgement instruction also utilizes an autoregressive manner but to judge whether a specific relation exists between objects, thereby simplifying the complexity of open-set relation prediction.
Next, we specify the two instructions, respectively.

\subsubsection{Generation Instruction.}
\label{sec:generation_inst}
For the generation instruction, we follow the instruction design used in open-set object recognition~\cite{yu2023towards}, utilizing ``What are the relations between $c_i$ and $c_j$?''.
Here, $c_i$ and $c_j$ respectively refers to the name of the subject and object.
We convert this instruction into features $F_{inst}^{gen} \in \mathbb{R}^{X^{gen} \times D}$ using the tokenizer, where $X^{gen}$ is the token number of the generation instruction.
We input the features of this generation instruction $F_{inst}^{gen}$ together with the subject-object pair features $F_{I}^{pair(i,j)}$ into the multimodal relation decoder $Dec(\cdot)$, predicting all possible relations in an autoregressive way, formulated as 
\begin{equation}
    r_{i,j} = Dec(Concat(F_{I}^{pair(i,j)}, F_{inst}^{gen})).
\end{equation}
If multiple relations are predicted, they are separated by the delimiter ``[SEP]''.

\subsubsection{Judgement Instruction.}
\label{sec:judgement_inst}

Unlike generation instruction, the judgement instruction guides relation decoder to judge, based on a given relation name, whether this relation exists between the subject and object.
For example, ``Please judge between $c_i$ and $c_j$ whether there is a relation $r_k$''.
In this case, we only need the multimodal relation decoder to answer ``Yes'' or ``No'' to determine the existence of this relation.
Note that inputting the complete judgement instruction for each relation into the decoder can be costly.
Therefore, we place the relation name at the end of the instruction.
During inference we divide the judgement instruction into two parts: the section before the relation name, transformed into $F_{inst}^{judge}$ through the tokenizer, and the relation name itself, processed into $F_{inst}^{rel}$.
Benefiting from the autoregressive manner for open-set relation prediction, we initially input the subject-object pair feature $F_{I}^{pair(i,j)}$ and $F_{inst}^{judge}$ into the multimodal relation decoder, formulated as
\begin{equation}
    F_{prefix}^{(i,j)} = Dec(Concat(F_{I}^{pair(i,j)}, F_{inst}^{judge})),
\end{equation}
which is then cached for subsequent calculations for each relation.
For each relation $r_k$, the multimodal relation decoder only needs to process $F_{prefix}$ and $F_{inst}^{rel(k)}$ to achieve relation prediction, formulated as:
\begin{equation}
    J_{i,j,k} = Dec(Concat(F_{prefix}^{(i,j)}, F_{inst}^{rel(k)})),
\end{equation}
where $J_{i,j,k}$ represents the judgement for $(o_i, r_k, o_j)$ triplet.
When $J_{i,j,k}$ is ``Yes'', it indicates that the relation $r_k$ exists between $o_i$ and $o_j$; otherwise, it does not exist.
Through this approach, we can maintain the same prediction time as with the generation instruction.

We perform the aforementioned process for all subject-object pairs that may have a relation, ultimately achieving open-set relation prediction.
For method using generation instruction, we denote it as OpenPSG-G, and for those using judgement instruction, as OpenPSG-J.
In next section, OpenPSG by default refers to the latter.

\subsection{Loss Function}
\label{sec:loss_function}
During the model training, there are two losses involved:
the binary cross-entropy loss $\mathcal{L}_{exist}$ for estimating the existence of a relation using the relation existence estimation query in the relation query transformer, and the cross-entropy loss $\mathcal{L}_{LM}$ consistent with language model training used by the multimodal relation decoder.
The total loss is:
$\mathcal{L} = \lambda \mathcal{L}_{exist} + \mathcal{L}_{LM}$,
where $\lambda$ is a weight factor.

%% file: sections/4_experiments.tex
\section{Experiments}
\label{sec:experiments}

\subsection{Datasets}

\noindent \textbf{Panoptic Scene Graph (PSG)} dataset~\cite{yang2022panoptic} is constructed based on the COCO dataset~\cite{lin2014microsoft, caesar2018coco}, consisting of 48,749 annotated images: 46,563 for training and 2,186 for testing. 
It encompasses 80 ``thing'' object categories~\cite{lin2014microsoft} and 53 ``stuff'' object categories~\cite{caesar2018coco}, as well as 56 relation categories.

\noindent \textbf{Visual Genome (VG)} dataset~\cite{krishna2017visual} is a widely used dataset in the SGG task.
To further validate our method, we follow previous works~\cite{yu2023visually, chen2023expanding} and test our method on the VG-150 variant, which contains 150 object categories and 50 relation categories.

\subsection{Tasks and metrics}

\noindent \textbf{Tasks.}
In both PSG and SGG, there are three distinct subtasks: Predicate Classification (PredCls), Scene Graph Classification (SGCls), and Scene Graph Detection (SGDet)~\cite{xu2017scene}.
In PredCls, the categories and locations of objects within the image are given, and only the relation categories between the objects need to be predicted.
SGCls requires predicting both the categories of objects and the relations between them, given the locations of objects within the image.
SGDet requires the simultaneous prediction of object categories, locations, and relations between objects, based on the given image.
In this paper, we focus on the PredCls and SGDet subtasks.
The PredCls excludes the influence of segmentation performance and only compares the relation prediction performance, while the SGDet considers the combined results for both object segmentation and relation prediction.

Furthermore, to validate our method's capability in open-set relation prediction, we divide the dataset into base relations and novel relations at a ratio of 7:3.
For the PSG dataset, please refer to the supplementary material for the specific division method.
For the division of the VG dataset, we follow the practice in previous works~\cite{yu2023visually, chen2023expanding}.
In the open-set scenarios, our model is trained with data only from base relations and tested on both base and novel relations.
It is noteworthy that the test sets for open-set and closed-set are same.

\noindent \textbf{Metrics.}
Following previous works~\cite{yang2022panoptic, zhou2023hilo}, we use Recall@K (R@K) and mean Recall@K (mR@K) as our evaluation metrics.
Additionally, in open-set scenarios, we also report the R@K and mR@K metrics for base and novel relations.

\subsection{Implementation details}
In our experiments, we utilize the pretrained OpenSeeD~\cite{zhang2023simple} as the open-set object segmenter.
The patch size $p$ of the patchify module is set to 8.
Within the relation query transformer, the length $E$ of the pair feature extraction query is 32, and the threshold $\theta$ used to filter subject-object pairs is set to 0.35.
In the multimodal relation decoder, we employ the decoder of BLIP-2~\cite{li2023blip}.
During model training, the weight factor $\lambda$ for the loss is set to 10.
We adopt the same data augmentation strategies as in previous methods~\cite{yang2022panoptic, zhou2023hilo}.
To train our model, we use the AdamW~\cite{loshchilov2017decoupled} optimizer with a learning rate of $1e^{-4}$ and a weight decay of $5e^{-2}$.
Our model is trained for a total of 12 epochs, reducing the learning rate to $1e^{-5}$ at the $8_{th}$ epoch.
The experimental platform uses four A100 GPUs.
Note that during training we freeze the parameters of the object segmenter and multimodal relation decoder but only train the proposed RelQ-Former.

\begin{table}[t]\small
    \centering
    \caption{Comparison between our OpenPSG and other methods on the PSG dataset in both closed-set and open-set scenarios.
    Our method shows superior performance compared to all previous methods.
    }
    \resizebox{\linewidth}{!}{
    \begin{tabular}{lcccccc}
    \toprule[1.5pt]
        ~ & \multicolumn{3}{c}{Predicate Classification} & \multicolumn{3}{c}{Scene Graph Detection} \\
        \cmidrule{2-7}
        Method & R/mR@20 & R/mR@50 & R/mR@100 & R/mR@20 & R/mR@50 & R/mR@100 \\
        \midrule[1pt]
        ~ & \multicolumn{6}{c}{Train on all relations (closed-set)} \\
        \cmidrule{2-7}
        IMP~\cite{xu2017scene} & 31.9/9.6~ & 36.8/10.9 & 38.9/11.6 & 16.5/6.5~ & 18.2/7.1~ & 18.6/7.2~ \\
        Motifs~\cite{zellers2018neural} & 44.9/20.2 & 50.4/22.1 & 52.4/22.9 & 20.0/9.1~ & 21.7/9.6~ & 22.0/9.7~ \\
        VCTree~\cite{tang2019learning} & 45.3/20.5 & 50.8/22.6 & 52.7/23.3 & 20.6/9.7~ & 22.1/10.2 & 22.5/10.2 \\
        GPSNet~\cite{lin2020gps} & 31.5/13.2 & 39.9/16.4 & 44.7/18.3 & 17.8/7.0~ & 19.6/7.5~ & 20.1/7.7~ \\
        PSGTR~\cite{yang2022panoptic} & ~--~/~--~ & ~--~/~--~ & ~--~/~--~ & 28.4/16.6 & 34.4/20.8 & 36.3/22.1 \\
        PSGFormer~\cite{yang2022panoptic} & ~--~/~--~ & ~--~/~--~ & ~--~/~--~ & 18.0/14.8 & 19.6/17.0 & 20.1/17.6 \\
        ADTrans~\cite{li2023panoptic} & ~~~--~/29.0 & ~~~--~/36.2 & ~~~--~/38.8 & 26.0/26.4 & 29.6/29.7 & 30.0/30.0 \\
        PairNet~\cite{wang2023pair} & ~--~/~--~ & ~--~/~--~ & ~--~/~--~ & 29.6/24.7 & 35.6/28.5 & 39.6/30.6 \\
        HiLo~\cite{zhou2023hilo} & ~--~/~--~ & ~--~/~--~ & ~--~/~--~ & 34.1/23.7 & 40.7/30.3 & 43.0/33.1 \\
	\textbf{OpenPSG} & \textbf{55.1}/\textbf{39.2} & \textbf{70.6}/\textbf{53.8} & \textbf{79.3}/\textbf{63.8} & \textbf{38.1}/\textbf{32.3} & \textbf{46.8}/\textbf{40.9} & \textbf{52.0}/\textbf{50.1} \\
        \midrule[1pt]
        ~ & \multicolumn{6}{c}{Train on base relations (open-set)} \\
        \cmidrule{2-7}
        \textbf{OpenPSG} & 45.1/29.1 & 55.5/38.7 & 61.5/46.0 & 25.9/20.9 & 31.6/24.0 & 36.7/25.4 \\
    \bottomrule[1.5pt]
    \end{tabular}
    }
    \label{tab:psg_results}
\end{table}

\subsection{Comparison to the state of the art}

\noindent \textbf{PSG dataset.}
Tab.~\ref{tab:psg_results} consists of two parts, comparing the performance of our method against previous methods in closed-set and open-set scenarios under the subtasks of predicate classification and scene graph detection.
For the first part, in the closed-set scenario, our method significantly surpasses previous methods.
For the predicate classification subtask, only methods that predict segmentation and relation separately are applicable~\cite{yang2022panoptic}, and our method achieves a substantial improvement compared to previous best results, for instance, a 26.6\% increase in R@100 and a 25.0\% increase in mR@100.
For scene graph detection, our method also achieves a significant increase of 9.0\% over the best previous method~\cite{zhou2023hilo} in R@100 and a 17.0\% increase in mR@100.
This indicates that in the closed-set scenario, our method demonstrates significant performance improvements.
For the second part, in the open-set scenario, we train the model only on base relations and test it on all relations.
Notably, for predicate classification, our method even outperforms previous methods trained on all relations, which demonstrates the superiority of our method.
For example, compared with previous best results, we achieve a 39\% improvement in R@100 and a 7.2\% improvement in mR@100.
For the scene graph detection subtask, our method with judgement instruction remains very competitive compared to \cite{yang2022panoptic} trained on all relations.

\noindent \textbf{VG dataset.}
To further validate our method on the VG dataset, we compare to two closed-set SGG methods~\cite{zellers2018neural, tang2019learning} and two recent open-set SGG works~\cite{yu2023visually, chen2023expanding}.
Since our method relies on an object segmentation model while \cite{zellers2018neural, tang2019learning, yu2023visually, chen2023expanding} rely on an object detector, for a fair comparison, we only present the results for predicate classification subtask.
Tab.~\ref{tab:vg_results} shows the results in both closed-set and open-set scenarios.
In the closed-set scenario, compared with previous closed-set methods~\cite{zellers2018neural, tang2019learning}, our method only performs a few points lower on R@50, yet yields significantly better results on the other metrics.
In addition, compared with previous best open-set methods~\cite{yu2023visually, chen2023expanding}, our method improves by 29.0\% in R@100 and by 9.5\% in mR@100.
In open-set scenario, our method improves upon\cite{yu2023visually, chen2023expanding} by 3.9\% in R@100 and by 5.4\% in mR@100.
These results demonstrate the effectiveness of our method in both closed-set and open-set relation prediction.

\begin{table}[t]\small
    \centering
    \caption{Comparison between our OpenPSG and other methods on the VG dataset in both closed-set and open-set scenarios with predicate classification subtask.}
    \resizebox{0.9\linewidth}{!}{
    \begin{tabular}{lcccc}
    \toprule[1.5pt]
        ~ & \multicolumn{2}{c}{Train on all relations (closed-set)} & \multicolumn{2}{c}{Train on base relations (open-set)} \\
        \cmidrule{2-5}
        Method & R/mR@50 & R/mR@100 & R/mR@50 & R/mR@100 \\
        \midrule[1pt]
        Motifs~\cite{zellers2018neural} & 65.2/15.9 & 67.1/17.2 & ~~~--~/~--~~ & ~~~--~/~--~~ \\
        VCTree~\cite{tang2019learning} & 66.4/16.8 & 68.1/19.4 & ~~~--~/~--~~ & ~~~--~/~--~~ \\
        Cacao+Epic~\cite{yu2023visually} & ~~~--~/39.0 & ~~~--~/40.8 & ~~~--~/16.5 & ~~~--~/21.8 \\
        OvSGTR~\cite{chen2023expanding} & 36.4/~--~~ & 42.4/~--~~ & 22.9/~--~~ & 26.7/~--~~ \\
        \textbf{OpenPSG} & ~\textbf{60.2}/\textbf{45.8} & ~\textbf{71.4}/\textbf{50.3} & ~\textbf{25.7}/\textbf{21.5} & ~\textbf{30.6}/\textbf{27.2} \\
    \bottomrule[1.5pt]
    \end{tabular}
    }
    \label{tab:vg_results}
\end{table}

\subsection{Ablation study}
To validate the effectiveness of each module in our method, we conduct ablation studies, using the model trained on all relations with judgement instruction for relation prediction as the baseline unless otherwise specified.
To eliminate interference, we validate the object segmenter and modules in RelQ-Former under closed-set settings, and test two types of instructions in the multimodal relation decoder under open-set settings.

\noindent \textbf{Different segmenters.}
Our method is based on a pretrained segmenter.
To verify the impact of different segmenters on model performance, we experiment with two options:
the closed-set Mask2Former~\cite{cheng2022masked} and the open-set OpenSeeD~\cite{zhang2023simple}. 
As shown in Tab.~\ref{tab:segmenter}, OpenSeeD outperforms Mask2Former on the Panoptic Quality (PQ)~\cite{kirillov2019panoptic} metric (55.1 \vs 51.7) in PSG test set in closed-set senario.
For the predicate classification subtask, the results show that using OpenSeeD is slightly better than using Mask2Former, with R@100 higher by 0.4\% and mR@100 by 0.1\%.
For the scene graph detection subtask, OpenSeeD is a better segmenter than Mask2Former, with an increase of 3.5\% in R@100 and 2.3\% in mR@100, due to its superior object segmentation capability (see PQ in Tab.~\ref{tab:segmenter}).

\begin{table}[t]\small
    \centering
    \caption{Ablation study of comparison of different segmenters.}
    \resizebox{\linewidth}{!}{
    \begin{tabular}{lccccccc}
    \toprule[1.5pt]
        ~ & ~ & \multicolumn{3}{c}{Predicate Classification} & \multicolumn{3}{c}{Scene Graph Detection} \\
        \cmidrule{2-8}
        Segmenter & PQ & R/mR@20 & R/mR@50 & R/mR@100 & R/mR@20 & R/mR@50 & R/mR@100 \\
        \midrule[1pt]
        OpenSeeD~\cite{zhang2023simple} & \textbf{55.1} & \textbf{55.1/39.2} & \textbf{70.6/53.8} & \textbf{79.3/63.8} &  \textbf{38.1/32.3} & \textbf{46.8/40.9} & \textbf{51.9/50.1} \\
        Mask2Former~\cite{cheng2022masked} & 51.7 & 54.8/39.1 & 70.4/52.9 & 78.9/63.7 &  36.1/30.3 & 44.8/38.4 & 48.4/47.8 \\
    \bottomrule[1.5pt]
    \end{tabular}
    }
    \label{tab:segmenter}
\end{table}

\noindent \textbf{Subject-object pair features extraction.}
To validate the effectiveness of the RelQ-Former in extracting subject-object pair features via the attention mechanism, we design an experiment by extracting subject-object pair features using mask pooling.
Specifically, for mask pooling, we derive object features by applying mask pooling to their respective positions in the visual features and then concatenate these to form the subject-object pair features.
As shown in Tab.~\ref{tab:ablation_rel_q_former}, it demonstrates that our attention mechanism outperforms mask pooling based pair feature extraction method by 5.2\% in R@100 and by 4.7\% in mR@100.
This indicates that our method using RelQ-Former can better focus the extracted features on the interactions between objects, thereby enhancing the performance for relation prediction.

\begin{table}[t]\small
    \centering
    \caption{Ablation study of comparison of different designs in RelQ-Former.}
    \resizebox{0.8\linewidth}{!}{
    \begin{tabular}{lcccccc}
    \toprule[1.5pt]
        ~ & \multicolumn{6}{c}{Predicate Classification} \\
        \cmidrule{2-7}
        Method & R@20 & mR@20 & R@50 & mR@50 & R@100 & mR@100 \\
        \midrule[1pt]
        RelQ-Former & \textbf{55.1} & \textbf{39.2} & \textbf{70.6} & \textbf{53.8} & \textbf{79.3} & \textbf{63.8} \\
        \hdashline
        Change to Mask Pooling & 51.5 & 36.3 & 65.9 & 49.3 & 74.1 & 59.1 \\
        Set $\theta=0$ in Selector & 56.8 & 40.7 & 70.9 & 54.0 & 79.6 & 63.9 \\
        Set $\lambda=0$ in Loss & 54.9 & 38.7 & 69.8 & 53.6 & 78.5 & 63.2 \\
    \bottomrule[1.5pt]
    \end{tabular}
    }
    \label{tab:ablation_rel_q_former}
\end{table}

\noindent \textbf{Selector in RelQ-Former.}
To validate the efficacy of the selector, we set $\theta$ to 0, meaning that during inference, we predict relations for all subject-object pairs.
As shown in Tab.~\ref{tab:ablation_rel_q_former}, we find that when we set $\theta$ to 0, the model performance is approximately the same.
However, under these conditions, the model takes 20 times longer to run on the whole PSG test set.
This indicates that our selector allows for a significant improvement in computational efficiency with a small sacrifice in performance.
For more details, please refer to supplementary material.

\begin{table}[t]\small
    \centering
    \caption{Ablation study of different instruction types in multimodal relation decoder under various base-to-novel relation ratios. OpenPSG-G: using generation instruction. OpenPSG-J: using judgement instruction.}
    \resizebox{0.9\linewidth}{!}{
    \begin{tabular}{ccccccc}
        \toprule[1.5pt]
        ~ & \multicolumn{3}{c}{OpenPSG-G} & \multicolumn{3}{c}{OpenPSG-J} \\
        \cmidrule{2-7}
        base:novel & R/mR@20 & R/mR@50 & R/mR@100 & R/mR@20 & R/mR@50 & R/mR@100 \\
        \midrule[1pt]
        7:3 & 40.1/24.5 & 49.4/32.4 & 57.1/36.8 &  45.1/29.1 & 55.5/38.7 & 61.5/46.0 \\
        6:4 & 34.9/21.3 & 46.3/29.2 & 52.9/32.0 &  40.6/25.1 & 53.9/35.3 & 57.0/43.5 \\
        5:5 & 25.4/14.0 & 38.6/23.2 & 40.1/24.2 &  36.4/19.2 & 48.9/29.3 & 50.3/40.8 \\
        4:6 & 19.8/11.2 & 32.9/14.3 & 33.2/18.8 &  29.8/17.3 & 40.9/21.6 & 44.3/34.7 \\
        3:7 & 13.0/10.7 & 17.6/13.0 & 18.5/14.4 &  22.8/15.7 & 28.3/19.7 & 35.1/23.7 \\
    \bottomrule[1.5pt]
    \end{tabular}
    }
    \label{tab:base_novel_ratio}
\end{table}

\noindent \textbf{Relation Existence Loss.}
To evaluate the impact of the relation existence estimation loss (Sec.~\ref{sec:loss_function}) on the model, we set the training $\lambda$ to 0, thereby removing this loss for model training.
We discover that this loss not only aids in training a relation existence classifier, but also has beneficial effect on the model.
As shown in Tab.~\ref{tab:ablation_rel_q_former}, setting $\lambda$ to 0 leads to a decrease in R@100 by 0.8\% and in mR@100 by 0.6\%.

\noindent \textbf{Analysis of instruction types in multimodal relation decoder.}
To further analyze the two types of instructions, generation instruction and judgement instruction of open-set relation prediction in our multimodal relation decoder (Sec.~\ref{sec:rel_decoder}), we conduct a series of experiments by adjusting the proportion of novel relation categories, conducting tests with base:novel ratios of 7:3, 6:4, 5:5, 4:6, and 3:7, to validate the performance, and results shown in Tab.~\ref{tab:base_novel_ratio}.
First, under the same base:novel ratio, the method using judgement instruction (OpenPSG-J) consistently outperforms the one using generation instruction (OpenPSG-G).
For example, at a base:novel ratio of 7:3, OpenPSG-J is 4.4\% higher than OpenPSG-G in R@100, and is 9.2\% higher in mR@100.
Second, the results indicate that as the proportion of novel relations increases, the performance of both OpenPSG-G and OpenPSG-G gradually decreases.
For OpenPSG-G, R@100 decreases from 57.1\% at a base:novel ratio of 7:3 to 18.5\% at a base:novel ratio of 3:7, a drop of 38.6\%.
OpenPSG-J's R@100 decreases by 26.4\%, which is less than the decrease for OpenPSG-G, indicating that OpenPSG-J has a stronger capability for relation prediction in an open world.

\begin{figure}[t]
    \centering
    \includegraphics[width=1.0\linewidth]{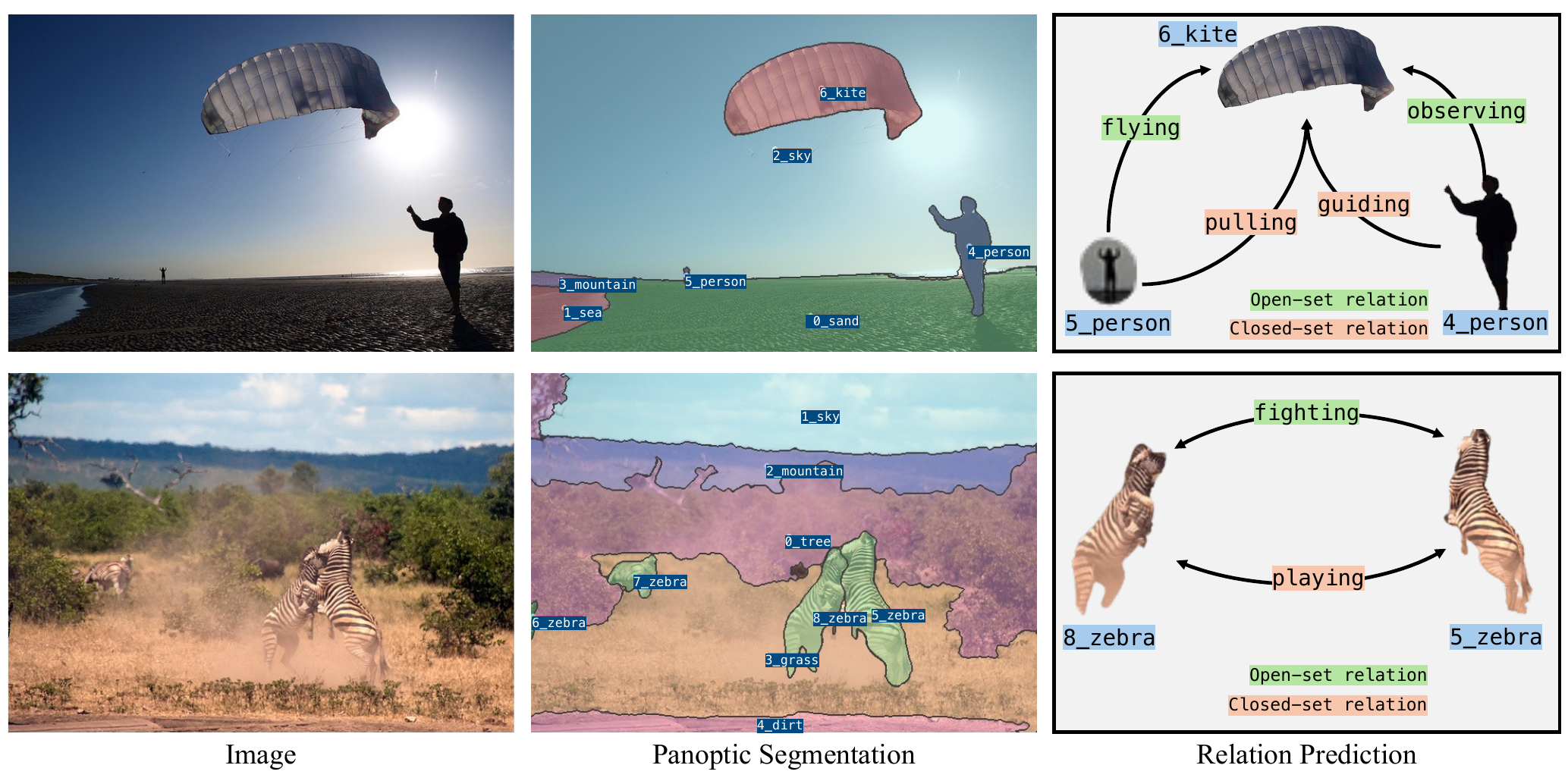}
    \caption{Visualization results produced by our OpenPSG.
    The left image is the input to our OpenPSG, the middle one displays the panoptic segmentation result, and the right one shows the predicted relations between objects.
    }
    \label{fig:vis2}
\end{figure}

\subsection{Visualization}
As shown in Fig.~\ref{fig:vis2}, our OpenPSG method can predict relations not defined in the dataset, such as ``\textit{flying}'', ``\textit{observing}'', and ``\textit{fighting}'', which qualitatively demonstrates the excellent open-set relation prediction capability of our method.

%% file: sections/5_conclusion.tex
\section{Conclusion}
\label{sec:conclusion}
In this paper, we propose the open-set PSG task and introduce the OpenPSG to accomplish open-set relation prediction.
With the help of large multimodal models, our method employs an autoregressive approach to predict open-set relations.
Additionally, we have developed a relation query transformer which contains pair feature extraction and relation existence estimation queries, one for extracting features of subject-object pairs, the other for predicting the existence of relations between them to filter out irrelevant pairs. 
Furthermore, we design generation and judgement instructions to enable the multimodal relation decoder to predict open-set relations.
Extensive experiments demonstrate that our method achieves excellent performance in open-set relation prediction.
In the future, we plan to employ model distillation to reduce the model size, thus enhancing the practicality in various real-world scenarios while ensuring its ability to predict open-set relations.

%% file: sections/6_appendix.tex
\appendix

\section*{Supplementary Material}

\noindent In this supplementary material, we provide the division method for base and novel relations in PSG dataset (Sec.~\ref{sec:division}), more experiments on other datasets (Sec.~\ref{sec:more_dataset_gqa}), additional experimental results (Sec.~\ref{sec:more_exp}), and instructions used in relation query transformer (Sec.~\ref{sec:instruction_rel_q_former}) and multimodal relation decoder (Sec.~\ref{sec:instruction_rel_decoder}) respectively.

\section{Division for base and novel relations in PSG dataset.}
\label{sec:division}
In the PSG dataset, there are a total of 56 predefined relations.
To test the open-set performance of our model, we divide the dataset into base and novel relations at a ratio of 7:3.
The novel relations, which make up 30\% of the total, while the rest are classified as base relations.
The detailed relations are as follows.
\begin{lstlisting}[language=Python]
base_relations = ["over", "in front of", "beside", "on", "in", "hanging from", "on back of", "going down", "painted on", "walking on", "running on", "crossing", "lying on", "sitting on", "jumping over", "jumping from", "holding", "carrying", "guiding", "kissing", "drinking", "feeding", "catching", "picking", "chasing", "climbing", "playing", "touching", "pulling", "opening", "talking to", "throwing", "driving", "riding", "driving on", "about to hit", "swinging", "entering", "exiting", "enclosing", "leaning on"]
novel_relations = ["attached to", "falling off", "walking on ", "standing on", "flying over", "wearing", "looking at", "eating", "biting", "playing with", "cleaning", "pushing", "cooking", "slicing", "parked on", "kicking", "existing"]
all_relations = base_relations + novel_relations
\end{lstlisting}

\section{More datasets}
\label{sec:more_dataset_gqa}
\textbf{GQA} dataset~\cite{hudson2019gqa} employs the same images as the VG dataset~\cite{krishna2017visual}, but includes more comprehensive annotations of objects and relations.
Following prior work~\cite{dong2022stacked}, we conduct our experiments using the GQA200 variant, which encompasses 200 object categories and 100 relation categories.

To further validate our method, we evaluate our method on the GQA200 in both closed-set and open-set SGG scenarios (Tab.~\ref{tab:rebuttal_gqa}).
Our method surpasses the previous best-performing method~\cite{dong2022stacked} by a sizeable margin (+5.6\% in R@100 and +1.3\% in mR@100).

\section{More experimental results}
\label{sec:more_exp}

\begin{figure}[t]
    \centering
    \includegraphics[width=0.6\linewidth]{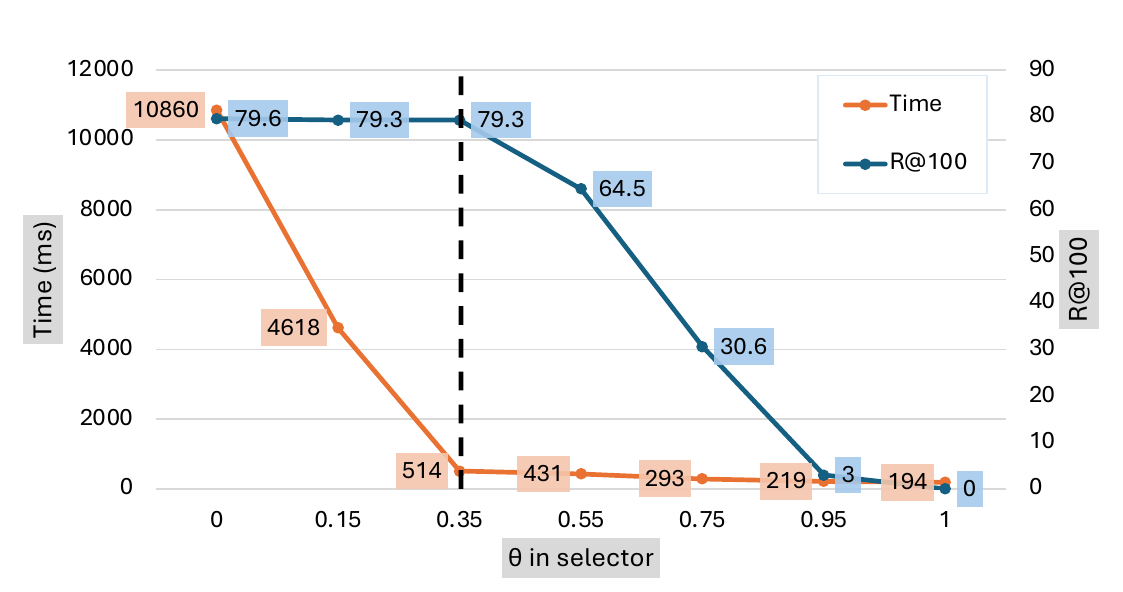}
    \caption{$\theta$ in the selector of RelQ-Former.}
    \label{fig:theta_time}
\end{figure}

\begin{table}[t]\small
    \centering
    \caption{Ablation study of RelQ-Former with different numbers of layers.}
    \resizebox{0.63\linewidth}{!}{
    \begin{tabular}{ccccccc}
        \toprule[1.5pt]
        ~ & \multicolumn{6}{c}{Predicate Classification} \\
        \cmidrule{2-7}
        Layer Num & R@20 & mR@20 & R@50 & mR@50 & R@100 & mR@100 \\
        \midrule[1pt]
        1 & 47.1 & 30.6 & 60.4 & 41.8 & 68.6 & 50.9 \\
        2 (Ours) & 55.1 & 39.2 & 70.6 & 53.8 & 79.3 & 63.8 \\
        4 & 55.1 & 39.5 & 70.7 & 54.0 & 79.8 & 64.1 \\
        6 & 54.2 & 38.5 & 69.4 & 52.8 & 78.3 & 62.8 \\
    \bottomrule[1.5pt]
    \end{tabular}
    }
    \label{tab:ablation_rel_q_former_layer}
\end{table}

\begin{table}[t]
    \centering
    \caption{Results on GQA: PredCls subtask in  closed- and open-set scenarios.}
    \resizebox{0.7\linewidth}{!}{
    \begin{tabular}{lcccc}
    \toprule[1.5pt]
        ~ & \multicolumn{2}{c}{Closed-set} & \multicolumn{2}{c}{Open-set (base:novel=7:3)} \\
        \cmidrule{2-5}
        Method & R/mR@50 & R/mR@100 & R/mR@50 & R/mR@100\\
        \midrule[1pt]
        SHA+GCL~\cite{dong2022stacked} & 41.0/42.7 & 42.7/44.5 & ~--~/~--~ & ~--~/~--~ \\
        OpenPSG & 46.2/44.1 & 48.3/45.8 & 23.6/22.7 & 26.2/25.0 \\
    \bottomrule[1.5pt]
    \end{tabular}
    }
    \label{tab:rebuttal_gqa}
\end{table}

\begin{table}[!ht]
    \centering
    \caption{Results on PSG: SGDet subtask in the open-set scenario.}
    \resizebox{0.8\linewidth}{!}{
    \begin{tabular}{lccc}
    \toprule[1.5pt]
        Method & R/mR@20 & R/mR@50 & R/mR@100\\
        \midrule[1pt]
        OpenPSG & 25.9/20.9 & 31.6/24.0 & 36.7/25.4 \\
        multi-division & \normalsize{26.4}\footnotesize{$\pm$0.7}\normalsize{/21.5}\footnotesize{$\pm$0.6} & \normalsize{31.4}\footnotesize{$\pm$0.4}\normalsize{/24.1}\footnotesize{$\pm$0.4} & \normalsize{36.5}\footnotesize{$\pm$0.4}\normalsize{/25.7}\footnotesize{$\pm$0.3} \\
        fuse G and J & 26.3/21.2 & 32.0/24.6 & 37.2/26.3 \\
    \bottomrule[1.5pt]
    \end{tabular}
    }
    \label{tab:rebuttal_psg}
\end{table}

\noindent \textbf{Effect of $\theta$ in the selector of RelQ-Former.}
To select the optimal $\theta$ to balance model performance and efficiency, we experiment with different $\theta$ parameters for the selector.
As shown in Fig.~\ref{fig:theta_time}, as $\theta$ increases from 0.0 to 0.35, the time required per image decreases rapidly, whereas there is no significant change as $\theta$ increases from 0.35 to 1.0.
At the same time, as $\theta$ rises from 0.0 to 0.35, there is only a minor decline in model performance; however, beyond 0.35, the performance of the model begins to decrease rapidly.
Therefore, to balance performance and efficiency, we select $\theta = 0.35$, thus ensuring a short average processing time for an image while maintaining high performance.

\noindent \textbf{Number of layers in RelQ-Former.}
To determine the number of layers for RelQ-Former, we conduct experiments with different numbers of layers, and the results are shown in Tab.~\ref{tab:ablation_rel_q_former_layer}.
When we increase the number of layers from 1 to 2, the model performance clearly improves, with R@100 increasing by 10.7\% and mR@100 by 12.9\%.
However, when the number of layers increases from 2 to 4, the model performance only sees a minimal improvement, with R@100 increasing by 0.5\% and mR@100 by 0.3\%.
Further increasing the number of layers from 4 to 6 leads to a slight decrease in model performance.
Considering that RelQ-Former requires a balanced layer number to extract features and learn relations without overwhelming computation for all possible subject-object pairs, i.e., $N \times (N-1)$ pairs.
Therefore, a layer number of 2 is the optimal choice for balancing performance and efficiency.

\noindent \textbf{Impact of different data divisions on the results.}
We conduct 5 random divisions of the base and novel classes in the PSG dataset and show that the results (mean and variance in Tab.~\ref{tab:rebuttal_psg}: multi-division) are rather stable, attesting to the robustness of our method.

\noindent \textbf{Fusion of generation and judgement instructions.}
Judgement instruction is used by default, while generation instruction is a variant inspired by~\cite{zhang2023simple} for comparison.
They are not combined in the paper, but we can simply fuse their outputs using the inference fusion method in \cite{zhou2023hilo, zhou2023vlprompt}, the results in Tab.~\ref{tab:rebuttal_psg} (fuse G and J) show slight improvement (+0.5\% in R@100) but also come with double computation cost.

\section{Instructions used in Relation Query Transformer}
\label{sec:instruction_rel_q_former}
We introduce the instructions used in the relation query former (Sec.~4.2).
To prevent the model from overfitting to a particular type of instruction during training, following ~\cite{liu2024visual, dai2024instructblip, yu2023towards}, we design 10 different variations for each category of instruction.
During training, one is randomly selected from these 10 options for use, whereas for inference, only the first one is chosen for evaluation.

\subsection{Instructions for Pair Feature Extraction Query.}

\begin{lstlisting}[language=Python]
# {subject} is the category name of subject.
# {object} is the category name of object.
instructions_for_feat_query = [
    "Please extract features for the {subject}-{object} pair based on the whole visual features of the image and the masks of the {subject} and {object}.",
    "Based on the image's holistic visual features and the masks of both {subject} and {object}, please derive the features of the {subject}-{object} pair.",
    "Utilizing the total visual features of the image, along with the {subject} and {object} masks, please identify the features of the {subject}-{object} pair.",
    "By considering the comprehensive visual features of the image and the respective masks of the {subject} and {object}, please isolate the features specific to the {subject}-{object} pair.",
    "Taking into account the global visual features of the image and the masks designated for the {subject} and {object}, please extract the particular features of the {subject}-{object} pair.",
    "Leveraging the entire visual features of the image as well as the masks for the {subject} and {object}, please delineate the features corresponding to the {subject}-{object} pair.",
    "Drawing on the overall visual features of the image and the masks of the {subject} and {object}, please ascertain the features for the {subject}-{object} pair.",
    "By harnessing the full visual features of the image along with the masks of the {subject} and {object}, please identify the distinctive features of the {subject}-{object} pair.",
    "With reference to the comprehensive visual features of the image and the masks for the {subject} and {object}, please extract the respective features of the {subject}-{object} pair.",
    "Considering the total visual features of the image and the defined masks for the {subject} and {object}, please determine the specific features of the {subject}-{object} pair.",
]
\end{lstlisting}

\subsection{Instructions for Relation Existence Estimation Query.}

\begin{lstlisting}[language=Python]
# {subject} is the category name of subject.
# {object} is the category name of object.
instructions_for_exist_query = [
    "Based on the visual features of the entire image and the masks for the {subject} and {object}, estimate whether there is a relation between the {subject} and {object}.",
    "Considering the holistic visual features of the image and the masks of the {subject} and {object}, determine whether a relation exists between the two entities.",
    "Utilizing the complete visual features of the image along with the {subject} and {object} masks, assess whether there is a relation between the {subject} and {object}.",
    "Given the overall visual features of the image and the masks for the {subject} and {object}, evaluate whether a relation exists between the {subject} and {object}.",
    "By analyzing the entire visual features of the image and the masks of the {subject} and {object}, ascertain whether there is a relation between the {subject} and {object}.",
    "With the comprehensive visual features of the image and the masks for both {subject} and {object}, deduce whether there is a relation between the {subject} and {object}.",
    "Reflecting on the total visual features of the image and the masks applied to the {subject} and {object}, gauge whether there is a relation between the {subject} and {object}.",
    "Considering the full visual features of the image and the masks of the {subject} and {object}, infer whether a relation exists between the {subject} and {object}.",
    "Leveraging the overall visual features of the image and the masks designated for the {subject} and {object}, identify whether there is a relation between the {subject} and {object}.",
    "By examining the entire visual features of the image and the masks for the {subject} and {object}, predict whether there is a relation between the {subject} and {object}.",
]
\end{lstlisting}

\section{Instructions used in Multimodal Relation Decoder}
\label{sec:instruction_rel_decoder}
We introduce the instructions used in the multimodal relation decoder (Sec.~4.3).
Same as Sec.~\ref{sec:instruction_rel_q_former}, to avoid model overfitting to a particular type of instruction during training, we design 10 different variations for each type of instruction.
During training, one is randomly selected from these 10 options for use, whereas for inference, only the first one is chosen for evaluation [39].

\subsection{Generation Instructions.}

\begin{lstlisting}[language=Python]
# {subject} is the category name of subject.
# {object} is the category name of object.
instruction_for_generation = [
    "Please determine what the relation is between {subject} and {object}.",
    "Please ascertain the relation between the {subject} and {object}.",
    "Please identify what relations exists between the {subject} and {object}.",
    "Please decide what kind of relation is present between the {subject} and the {object}.",
    "Please deduce the relation between the {subject} and the {object}.",
    "Please establish what the relation is between the {subject} and {object}.",
    "Please clarify the relation between the {subject} and {object}.",
    "Please determine the type of relation existing between the {subject} and the {object}.",
    "Please pinpoint the kind of relation between the {subject} and {object}.",
    "Please evaluate what the relation is between the {subject} and the {object}.",
]
\end{lstlisting}

\subsection{Judgement Instructions.}

\begin{lstlisting}[language=Python]
# {subject} is the category name of subject.
# {object} is the category name of object.
# {relation} is the category name of relation.
instructions_for_judgement = [
    "Please judge between {subject} and {object} whether there is a relation {relation}.",
    "Please determine if there exists a relation between the {subject} and {object}, termed {relation}.",
    "Please ascertain whether there is a relation between the {subject} and {object}, identified as {relation}.",
    "Please evaluate if a relation between the {subject} and {object} can be classified as {relation}.",
    "Please judge whether there is a relation between the {subject} and {object} referred to as {relation}.",
    "Please decide if there is a relation between the {subject} and {object} denoted as {relation}.",
    "Please establish whether there is a relation between the {subject} and {object}, described as {relation}.",
    "Please conclude whether a relation exists between the {subject} and {object}, designated as {relation}.",
    "Please investigate whether there is a relation between the {subject} and {object}, recognized as {relation}.",
    "Please analyze if there exists a relation between the {subject} and {object}, characterized as {relation}.",
]
\end{lstlisting}

%% file: main.bbl
\begin{thebibliography}{10}
\providecommand{\url}[1]{\texttt{#1}}
\providecommand{\urlprefix}{URL }
\providecommand{\doi}[1]{https://doi.org/#1}

\bibitem{achiam2023gpt}
Achiam, J., Adler, S., Agarwal, S., Ahmad, L., Akkaya, I., Aleman, F.L., Almeida, D., Altenschmidt, J., Altman, S., Anadkat, S., et~al.: Gpt-4 technical report. arXiv preprint arXiv:2303.08774  (2023)

\bibitem{caesar2018coco}
Caesar, H., Uijlings, J., Ferrari, V.: Coco-stuff: Thing and stuff classes in context. In: CVPR (2018)

\bibitem{carion2020end}
Carion, N., Massa, F., Synnaeve, G., Usunier, N., Kirillov, A., Zagoruyko, S.: End-to-end object detection with transformers. In: ECCV. pp. 213--229 (2020)

\bibitem{chen2020say}
Chen, S., Jin, Q., Wang, P., Wu, Q.: Say as you wish: Fine-grained control of image caption generation with abstract scene graphs. In: CVPR (2020)

\bibitem{chen2023expanding}
Chen, Z., Wu, J., Lei, Z., Zhang, Z., Chen, C.: Expanding scene graph boundaries: Fully open-vocabulary scene graph generation via visual-concept alignment and retention. arXiv preprint arXiv:2311.10988  (2023)

\bibitem{cheng2022masked}
Cheng, B., Misra, I., Schwing, A.G., Kirillov, A., Girdhar, R.: Masked-attention mask transformer for universal image segmentation. In: CVPR. pp. 1290--1299 (2022)

\bibitem{dai2015convolutional}
Dai, J., He, K., Sun, J.: Convolutional feature masking for joint object and stuff segmentation. In: Proceedings of the IEEE conference on computer vision and pattern recognition. pp. 3992--4000 (2015)

\bibitem{dai2024instructblip}
Dai, W., Li, J., Li, D., Tiong, A.M.H., Zhao, J., Wang, W., Li, B., Fung, P.N., Hoi, S.: Instructblip: Towards general-purpose vision-language models with instruction tuning. NeurIPS  (2024)

\bibitem{dong2022stacked}
Dong, X., Gan, T., Song, X., Wu, J., Cheng, Y., Nie, L.: Stacked hybrid-attention and group collaborative learning for unbiased scene graph generation. In: Proceedings of the IEEE/CVF Conference on Computer Vision and Pattern Recognition. pp. 19427--19436 (2022)

\bibitem{dosovitskiy2020image}
Dosovitskiy, A., Beyer, L., Kolesnikov, A., Weissenborn, D., Zhai, X., Unterthiner, T., Dehghani, M., Minderer, M., Heigold, G., Gelly, S., et~al.: An image is worth 16x16 words: Transformers for image recognition at scale. arXiv preprint arXiv:2010.11929  (2020)

\bibitem{du2022learning}
Du, Y., Wei, F., Zhang, Z., Shi, M., Gao, Y., Li, G.: Learning to prompt for open-vocabulary object detection with vision-language model. In: CVPR. pp. 14084--14093 (2022)

\bibitem{gao2018image}
Gao, L., Wang, B., Wang, W.: Image captioning with scene-graph based semantic concepts. In: ICMLC (2018)

\bibitem{ghiasi2022scaling}
Ghiasi, G., Gu, X., Cui, Y., Lin, T.Y.: Scaling open-vocabulary image segmentation with image-level labels. In: ECCV. pp. 540--557. Springer (2022)

\bibitem{he2022towards}
He, T., Gao, L., Song, J., Li, Y.F.: Towards open-vocabulary scene graph generation with prompt-based finetuning. In: ECCV (2022)

\bibitem{hildebrandt2020scene}
Hildebrandt, M., Li, H., Koner, R., Tresp, V., G{\"u}nnemann, S.: Scene graph reasoning for visual question answering. arXiv preprint arXiv:2007.01072  (2020)

\bibitem{hudson2019gqa}
Hudson, D.A., Manning, C.D.: Gqa: A new dataset for real-world visual reasoning and compositional question answering. In: Proceedings of the IEEE/CVF conference on computer vision and pattern recognition. pp. 6700--6709 (2019)

\bibitem{joshi2019bert}
Joshi, M., Levy, O., Weld, D.S., Zettlemoyer, L.: Bert for coreference resolution: Baselines and analysis. arXiv preprint arXiv:1908.09091  (2019)

\bibitem{kan2021zero}
Kan, X., Cui, H., Yang, C.: Zero-shot scene graph relation prediction through commonsense knowledge integration. In: Machine Learning and Knowledge Discovery in Databases. Research Track: European Conference, ECML PKDD 2021, Bilbao, Spain, September 13--17, 2021, Proceedings, Part II 21. pp. 466--482. Springer (2021)

\bibitem{kirillov2019panoptic}
Kirillov, A., He, K., Girshick, R., Rother, C., Doll{\'a}r, P.: Panoptic segmentation. In: Proceedings of the IEEE/CVF conference on computer vision and pattern recognition. pp. 9404--9413 (2019)

\bibitem{krishna2017visual}
Krishna, R., Zhu, Y., Groth, O., Johnson, J., Hata, K., Kravitz, J., Chen, S., Kalantidis, Y., Li, L.J., Shamma, D.A., et~al.: Visual genome: Connecting language and vision using crowdsourced dense image annotations. International journal of computer vision pp. 32--73 (2017)

\bibitem{li2023blip}
Li, J., Li, D., Savarese, S., Hoi, S.: Blip-2: Bootstrapping language-image pre-training with frozen image encoders and large language models. arXiv preprint arXiv:2301.12597  (2023)

\bibitem{li2023panoptic}
Li, L., Ji, W., Wu, Y., Li, M., Qin, Y., Wei, L., Zimmermann, R.: Panoptic scene graph generation with semantics-prototype learning. arXiv preprint arXiv:2307.15567  (2023)

\bibitem{liang2023open}
Liang, F., Wu, B., Dai, X., Li, K., Zhao, Y., Zhang, H., Zhang, P., Vajda, P., Marculescu, D.: Open-vocabulary semantic segmentation with mask-adapted clip. In: CVPR. pp. 7061--7070 (2023)

\bibitem{lin2022learning}
Lin, C., Sun, P., Jiang, Y., Luo, P., Qu, L., Haffari, G., Yuan, Z., Cai, J.: Learning object-language alignments for open-vocabulary object detection. arXiv preprint arXiv:2211.14843  (2022)

\bibitem{lin2014microsoft}
Lin, T.Y., Maire, M., Belongie, S., Hays, J., Perona, P., Ramanan, D., Doll{\'a}r, P., Zitnick, C.L.: Microsoft coco: Common objects in context. In: ECCV (2014)

\bibitem{lin2020gps}
Lin, X., Ding, C., Zeng, J., Tao, D.: Gps-net: Graph property sensing network for scene graph generation. In: CVPR (2020)

\bibitem{liu2023improved}
Liu, H., Li, C., Li, Y., Lee, Y.J.: Improved baselines with visual instruction tuning. arXiv preprint arXiv:2310.03744  (2023)

\bibitem{liu2024visual}
Liu, H., Li, C., Wu, Q., Lee, Y.J.: Visual instruction tuning. NeurIPS  \textbf{36} (2024)

\bibitem{loshchilov2017decoupled}
Loshchilov, I., Hutter, F.: Decoupled weight decay regularization. In: ICLR (2019)

\bibitem{radford2021learning}
Radford, A., Kim, J.W., Hallacy, C., Ramesh, A., Goh, G., Agarwal, S., Sastry, G., Askell, A., Mishkin, P., Clark, J., et~al.: Learning transferable visual models from natural language supervision. In: ICML. pp. 8748--8763 (2021)

\bibitem{raffel2020exploring}
Raffel, C., Shazeer, N., Roberts, A., Lee, K., Narang, S., Matena, M., Zhou, Y., Li, W., Liu, P.J.: Exploring the limits of transfer learning with a unified text-to-text transformer. The Journal of Machine Learning Research  \textbf{21}(1),  5485--5551 (2020)

\bibitem{singh2023scene}
Singh, K.P., Salvador, J., Weihs, L., Kembhavi, A.: Scene graph contrastive learning for embodied navigation. In: ICCV (2023)

\bibitem{tang2019learning}
Tang, K., Zhang, H., Wu, B., Luo, W., Liu, W.: Learning to compose dynamic tree structures for visual contexts. In: CVPR (2019)

\bibitem{touvron2023llama}
Touvron, H., Lavril, T., Izacard, G., Martinet, X., Lachaux, M.A., Lacroix, T., Rozi{\`e}re, B., Goyal, N., Hambro, E., Azhar, F., et~al.: Llama: Open and efficient foundation language models. arXiv preprint arXiv:2302.13971  (2023)

\bibitem{wang2020language}
Wang, C., Liu, X., Song, D.: Language models are open knowledge graphs. arXiv preprint arXiv:2010.11967  (2020)

\bibitem{wang2023pair}
Wang, J., Wen, Z., Li, X., Guo, Z., Yang, J., Liu, Z.: Pair then relation: Pair-net for panoptic scene graph generation. arXiv preprint arXiv:2307.08699  (2023)

\bibitem{xu2017scene}
Xu, D., Zhu, Y., Choy, C.B., Fei-Fei, L.: Scene graph generation by iterative message passing. In: CVPR (2017)

\bibitem{yang2023focusing}
Yang, J., Wang, C., Liu, Z., Wu, J., Wang, D., Yang, L., Cao, X.: Focusing on flexible masks: A novel framework for panoptic scene graph generation with relation constraints. In: ACM MM. pp. 4209--4218 (2023)

\bibitem{yang2022panoptic}
Yang, J., Ang, Y.Z., Guo, Z., Zhou, K., Zhang, W., Liu, Z.: Panoptic scene graph generation. In: ECCV. pp. 178--196 (2022)

\bibitem{yao2023detclipv2}
Yao, L., Han, J., Liang, X., Xu, D., Zhang, W., Li, Z., Xu, H.: Detclipv2: Scalable open-vocabulary object detection pre-training via word-region alignment. In: CVPR. pp. 23497--23506 (2023)

\bibitem{yu2023visually}
Yu, Q., Li, J., Wu, Y., Tang, S., Ji, W., Zhuang, Y.: Visually-prompted language model for fine-grained scene graph generation in an open world. arXiv preprint arXiv:2303.13233  (2023)

\bibitem{yu2023towards}
Yu, Q., Shen, X., Chen, L.C.: Towards open-ended visual recognition with large language model. arXiv preprint arXiv:2311.08400  (2023)

\bibitem{yu2022zero}
Yu, X., Chen, R., Li, J., Sun, J., Yuan, S., Ji, H., Lu, X., Wu, C.: Zero-shot scene graph generation with knowledge graph completion. In: 2022 IEEE International Conference on Multimedia and Expo (ICME). pp.~1--6. IEEE (2022)

\bibitem{yue2023object}
Yue, K., Chen, B.C., Geiping, J., Li, H., Goldstein, T., Lim, S.N.: Object recognition as next token prediction. arXiv preprint arXiv:2312.02142  (2023)

\bibitem{zang2022open}
Zang, Y., Li, W., Zhou, K., Huang, C., Loy, C.C.: Open-vocabulary detr with conditional matching. In: ECCV. pp. 106--122. Springer (2022)

\bibitem{zellers2018neural}
Zellers, R., Yatskar, M., Thomson, S., Choi, Y.: Neural motifs: Scene graph parsing with global context. In: CVPR (2018)

\bibitem{zhang2023simple}
Zhang, H., Li, F., Zou, X., Liu, S., Li, C., Yang, J., Zhang, L.: A simple framework for open-vocabulary segmentation and detection. In: ICCV. pp. 1020--1031 (2023)

\bibitem{zhang2023learning}
Zhang, Y., Pan, Y., Yao, T., Huang, R., Mei, T., Chen, C.W.: Learning to generate language-supervised and open-vocabulary scene graph using pre-trained visual-semantic space. In: Proceedings of the IEEE/CVF Conference on Computer Vision and Pattern Recognition. pp. 2915--2924 (2023)

\bibitem{zhao2023textpsg}
Zhao, C., Shen, Y., Chen, Z., Ding, M., Gan, C.: Textpsg: Panoptic scene graph generation from textual descriptions. In: ICCV. pp. 2839--2850 (2023)

\bibitem{zhou2023hilo}
Zhou, Z., Shi, M., Caesar, H.: Hilo: Exploiting high low frequency relations for unbiased panoptic scene graph generation. arXiv preprint arXiv:2303.15994  (2023)

\bibitem{zhou2023vlprompt}
Zhou, Z., Shi, M., Caesar, H.: Vlprompt: Vision-language prompting for panoptic scene graph generation. arXiv preprint arXiv:2311.16492  (2023)

\bibitem{zhu2023minigpt}
Zhu, D., Chen, J., Shen, X., Li, X., Elhoseiny, M.: Minigpt-4: Enhancing vision-language understanding with advanced large language models. arXiv preprint arXiv:2304.10592  (2023)

\end{thebibliography}
